\documentclass[letterpaper, 10 pt, conference]{ieeeconf}  % ICRA format

\IEEEoverridecommandlockouts                              % This command is only
\let\labelindent\relax

\usepackage{times}
\usepackage{amsmath,amsfonts}
\usepackage{algorithmic}
\usepackage{algorithm}
\usepackage{array}
\usepackage{balance}
\usepackage[dvipsnames,table]{xcolor}
\usepackage{textcomp}
\usepackage{stfloats}
\usepackage{url}
\usepackage{verbatim}
\usepackage{graphicx}
\usepackage{todonotes}
\usepackage{booktabs} %needed for nice tables
\usepackage{multirow}
\usepackage{multicol}
\usepackage{bbm} % indicator function
\usepackage{mathtools}
\usepackage{pgf} % for calculating the values for gradient
\usepackage{etoolbox} % robustify command
\usepackage{enumitem} % to modify itemize environments
\usepackage{amssymb}

\usepackage{hyperref} % to show links to the references, and also back to where the reference was cited
\usepackage{lipsum} % Dummy text (lorem ipsum)
\usepackage{siunitx} % SI units
\usepackage{mathtools}
\usepackage{xspace}
\usepackage[acronym, nonumberlist, nogroupskip]{glossaries-extra}

\usepackage{pbox}
\usepackage{threeparttable}
\usepackage{tablefootnote}

\usepackage{tikz}

\usepackage[normalem]{ulem}
\usepackage{gensymb}

\definecolor{MK_Two_One}{RGB}{178,24,43} % Nice red
\definecolor{MK_Two_Two}{RGB}{239,138,98}
\definecolor{MK_Two_Three}{RGB}{253,219,199}
\definecolor{MK_Two_Four}{RGB}{209,229,240}
\definecolor{MK_Two_Five}{RGB}{103,169,207}
\definecolor{MK_Two_Six}{RGB}{33,102,172} % Nice blue

\hypersetup{
colorlinks=true
,linkcolor=black
,citecolor=black
,filecolor=MK_Two_Six
,urlcolor= MK_Two_Six
,menucolor=MK_Two_Five
,runcolor=MK_Two_Four
,linkbordercolor=MK_Two_One
,citebordercolor=MK_Two_Two
,filebordercolor=MK_Two_Three
,urlbordercolor=MK_Two_Six
,menubordercolor=MK_Two_Five
,runbordercolor=MK_Two_Four
}

\usepackage{subcaption}
\definecolor{high}{RGB}{116, 173, 209}  % the color for the highest number in your data set
\definecolor{low}{RGB}{244, 109, 67}  % the color for the lowest number in your data set
\def\secref#1{Sec.~\ref{#1}}

\def\figref#1{Fig.~\ref{#1}}
\def\tabref#1{Tab.~\ref{#1}}
\def\eqref#1{Eq.~(\ref{#1})}

\newcommand{\etal}{\textit{et al.}}

\usepackage{color}
\definecolor{ColorLightCyan}{rgb}{0.88,1,1}
\definecolor{ColorLightTurquoise}{rgb}{0.5, 1, 0.8}
\definecolor{ColorOrange}{rgb}{1.0, 0.7, 0.0}
\definecolor{ColorTrajBlue}{rgb}{0,0.5,1.0}
\definecolor{ColorTrajOrange}{rgb}{0.87,0.56,0.01}

\usepackage{tikz}

\newcolumntype{L}[1]{>{\raggedright\let\newline\\\arraybackslash\hspace{0pt}}m{#1}}
\newcolumntype{C}[1]{>{\centering\let\newline\\\arraybackslash\hspace{0pt}}m{#1}}
\newcolumntype{R}[1]{>{\raggedleft\let\newline\\\arraybackslash\hspace{0pt}}m{#1}}

\robustify{\gls}

\glssetcategoryattribute{acronym}{indexonlyfirst}{true}

\setabbreviationstyle[acronym]{long-short}

\glssetcategoryattribute{acronym}{nohyperfirst}{true}

\newacronym{ape}{APE}{Absolute Pose Error}
\newacronym{apf}{APF}{Artificial Potential Field}
\newacronym{als}{ALS}{Aerial Laser Scanning}

\newacronym{chomp}{CHOMP}{Covariant Hamiltonian Optimization for Motion Planning}
\newacronym{cpm}{CPM}{Camera Performance Model}
\newacronym{cvar}{CVaR}{Conditional Value-at-Risk}

\newacronym{darpa}{DARPA}{Defense Advanced Research Projects Agency}
\newacronym{dbh}{DBH}{Diameter at Breast Height}
\newacronym{dof}{DoF}{Degrees of Freedom}
\newacronym{ddp}{DDP}{Differential Dynamic Programming}
\newacronym{dss}{DSS}{Decision Support System}

\newacronym{fov}{FoV}{Field-of-View}
\newacronym{fmm}{FMM}{Fast Marching Method}

\newacronym{gd}{GD}{Gradient Descent}
\newacronym{gdf}{GDF}{Geodesic Distance Field}
\newacronym{gn}{GN}{Gauss-Newton}
\newacronym{gps}{GPS}{Global Positioning System}
\newacronym{gpmp}{GPMP}{Gaussian Process Motion Planner}
\newacronym{gpu}{GPU}{Graphics Processing Unit}

\newacronym{icp}{ICP}{Iterative Closest Point}
\newacronym{imu}{IMU}{Inertial Measurement Unit}
\newacronym{isam}{iSAM}{Incremental Smoothing and Mapping}
\newacronym{isp}{ISP}{Image Signal Processor}

\newacronym{komo}{KOMO}{k-Order Motion Optimization}

\newacronym{lagr}{LAGR}{Learning Applied to Ground Vehicles}
\newacronym{lhs}{LHS}{Left-hand Side}
\newacronym{lidar}{LiDAR}{Light Detection and Ranging}
\newacronym{lm}{LM}{Levenberg-Marquardt}
\newacronym{ls}{LS}{Least Squares}

\newacronym{map}{MAP}{Maximum A Posteriori}
\newacronym{mdbi}{MDBI}{Mean Distance Between Interventions}
\newacronym{mpc}{MPC}{Model Predictive Control}
\newacronym{mtbi}{MTBI}{Mean Time Between Interventions}

\newacronym{nls}{NLS}{Non-linear Least Squares}

\newacronym{pnp}{PnP}{Perspective-n-Points}
\newacronym{prm}{PRM}{Probabilistic Roadmap}
\newacronym{pte}{PTE}{Pose Tracking Error}

\newacronym{rgb}{RGB}{Red-Green-Blue}
\newacronym[\glsshortpluralkey=RMPs,\glslongpluralkey=Riemannian Motion Policies]{rmp}{RMP}{Riemannian Motion Policy}
\newacronym{rmse}{RMSE}{Root Mean Square Error}
\newacronym{rrt}{RRT}{Rapidly-exploring Random Tree}
\newacronym{ros}{ROS}{Robot Operating System}

\newacronym{sam}{SAM}{Smoothing and Mapping}
\newacronym{sdf}{SDF}{Signed Distance Field}
\newacronym{sfm}{SfM}{Structure-from-Motion}
\newacronym{slam}{SLAM}{Simultaneous Localization And Mapping}
\newacronym{sgd}{SGD}{Stochastic Gradient Descent}

\newacronym{to}{TO}{Trajectory Optimization}
\newacronym{tof}{ToF}{Time-of-Flight}
\newacronym{tls}{TLS}{Terrestrial Laser Scanning}

\newacronym{ugv}{UGV}{Unmanned Ground Vehicle}
\newacronym{urdf}{URDF}{Unified Robot Description Format}

\newacronym{vtr}{VT\&R}{Visual Teach and Repeat}
\newacronym{vit}{ViT}{Visual Transformer}

\newcommand{\subarrow}[1]{
	\mathord{
		\renewcommand{\arraystretch}{0}
		\begin{array}[t]{@{}c@{}l@{}}
			#1\\[2pt]
			\hspace{-2pt}\scriptstyle\longrightarrow
		\end{array}
		\kern\scriptspace
	}
}

\usepackage{fancyhdr}
\fancypagestyle{withfooter}{
  
  \fancyfoot[C]{\footnotesize Accepted to the IEEE ICRA Workshop on Field Robotics 2024}
}

\begin{document}
% --------------------------------------------------
% Title
\title{ \LARGE \bf Autonomous Forest Inventory with Legged Robots:\\System Design and Field Deployment
}

% --------------------------------------------------
% Authors
\author{Matias Mattamala$^{1}$, Nived Chebrolu$^{1}$, Benoit Casseau$^{1}$, Leonard Freißmuth$^{1,3}$, \\ Jonas Frey$^{2,4}$, Turcan Tuna$^{2}$, Marco Hutter$^{2}$, Maurice Fallon$^{1}$
\thanks{$^{1}$The authors are with the Oxford Robotics Institute at the University of
Oxford, UK. {\tt\small \{matias, nived, benoit, mfallon\}@robots.ox.ac.uk}
}
\thanks{$^{2}$The authors are with the Robotic Systems Lab, ETH Zurich. {\tt\small \{jonfrey, tutuna, mahutter\}@ethz.ch}
}
\thanks{$^{3}$The author is also with the Technical University of Munich, Germany. \texttt{\{l.freissmuth\}@tum.de}.
}
\thanks{$^{4}$The author is also with the Autonomous Learning Group, Max Planck Institute for Intelligent Systems.
}
}

% Title
\maketitle

% This makes sure that repeated author names in subsequent papers shown in the
% references are not dashed
\bstctlcite{IEEEexample:BSTcontrol}

\begin{abstract}
We present a solution for autonomous forest inventory with a legged robotic platform. Compared to their wheeled and aerial counterparts, legged platforms offer an attractive balance of endurance and low soil impact for forest applications. In this paper, we present the complete system architecture of our forest inventory solution which includes state estimation, navigation, mission planning, and real-time tree segmentation and trait estimation. We present preliminary results for three campaigns in forests in Finland and the UK and summarize the main outcomes, lessons, and challenges. Our UK experiment at the Forest of Dean with the ANYmal D legged platform, achieved an autonomous survey of a \SI{0.96}{\hectare} plot in \SI{20}{\minute}, identifying over 100 trees with typical DBH accuracy of \SI{2}{\centi\meter}.
\end{abstract}

% Field robotics WS add-on
\thispagestyle{withfooter}
\pagestyle{withfooter}

%\IEEEpeerreviewmaketitle
\section{Introduction}
\label{sec:intro}
Systematically measuring tree attributes such as the tree height, \gls{dbh}, and canopy volume, is important for modern forestry~\cite{Brack2001}. This process, known as forest inventory, has traditionally involved manual tree measurement using measuring tapes and calipers and also handheld laser rangefinders~\cite{Hamilton1988}. Nowadays, remote sensing techniques such as \gls{tls}~\cite{Dubayah2000} and aerial photograph~\cite{Hall2003} have massively improved the accuracy and scale at which forest inventories are built~\cite{Liang2016}.

Robotics has the potential to scale this process even further. Technologies such as \gls{slam} have been used for handheld forest mapping~\cite{Miettinen2007, Tremblay2020}. Fixed-wing aircrafts~\cite{Koh2012} and hexacopters~\cite{Almeida2020} have enabled autonomous data collection above the tree canopy where there is no risk of collision. Meanwhile small-scale drones~\cite{Chen2020, Liu2022} and wheeled platforms~\cite{Pierzchala2018} have been recently used for under-canopy forest inventory missions. Legged robots potentially provide significant advantages to aerial and wheeled platforms in terms of mobility, endurance, and lower soil impact~\cite{Todd1985}. However, robots have not been used for forestry tasks apart from tree harvesting~\cite{Jelavic2021}.

In this work, we study the application of legged platforms for autonomous forest inventory tasks. Our solution combines the advanced mobility skills developed by the robotics research community~\cite{Miki2022a}, a robust LiDAR-based mapping system with a multi-level autonomous mission planning architecture. Our solution leverages these components to provide a consistent representation of the environment, while also enabling online processing and concurrent estimation of the forest inventory.

\begin{figure}[t]
  \centering
  \includegraphics[width=\linewidth]{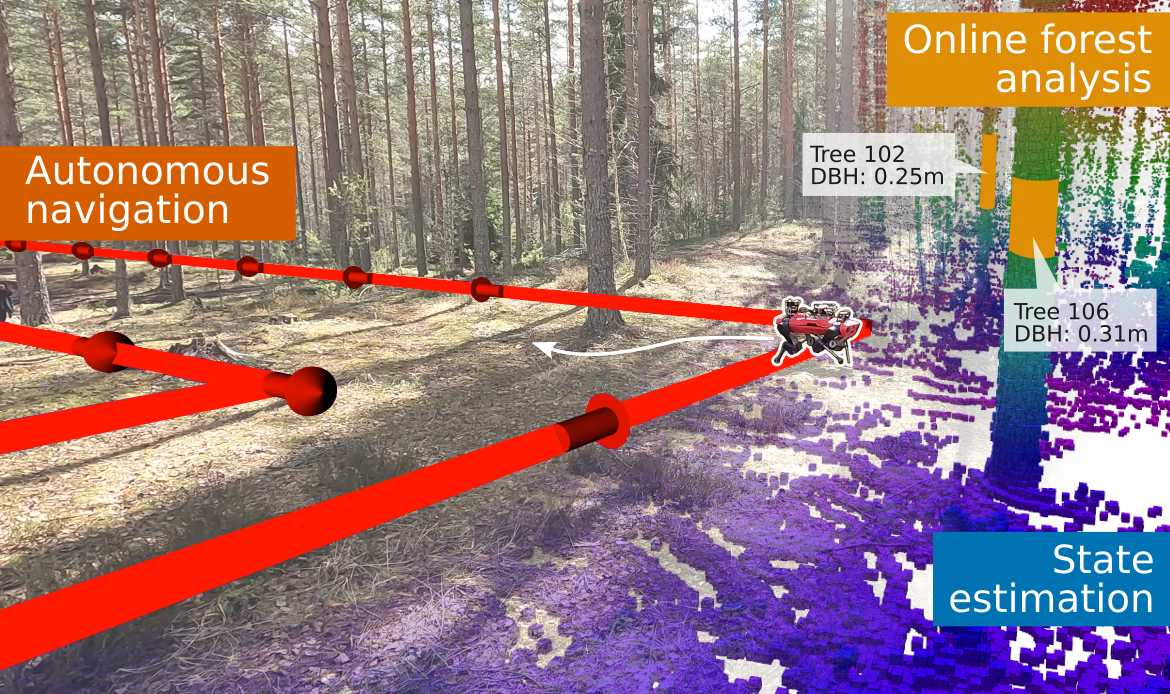}
  \caption{\small{We present an autonomous forest inventory solution with legged platforms. Our system aims to autonomously drive a legged platform in an unknown forest plot, while recording environmental data used for tree segmentation and trait estimation.}}
  \label{fig:deployments}
\end{figure}

This paper presents ongoing work, reporting autonomous deployments executed in May 2023, October 2023, and February 2024 using our solution. Our experiments have been carried out in various forests in Finland and the UK, as shown in \figref{fig:deployments}, and reflect the evolving design decisions we have made after each deployment. We discuss the findings and lessons learned, as well as the main challenges towards future autonomous forest inventory missions in the context of the DigiForest EU Horizon project\footnote{\url{https://www.digiforest.eu}}.
%
% --------------------------------------------------
% --------------------------------------------------

\section{Related Work}
\label{sec:related-work}

\subsection{Robotics-enabled forestry}
Robotics has the potential to scale up forest mensuration through technologies and platforms that can automate intensive manual labor. \gls{slam} can accelerate ground-based forest mensuration by integrating LiDAR scans from a mobile sensing payload. The produced maps can be post-processed to extract the desired attributes, such as the tree positions and \gls{dbh}~\cite{Miettinen2007,Tang2015,Tremblay2020}.
The incremental nature of \gls{slam} and the data acquisition process has motivated the concurrent construction of maps and inventories. Chen~\etal{}~\cite{Chen2020} and Proudman~\etal{}~\cite{Proudman2022}, demonstrated real-time tree detection and \gls{dbh} estimation using small-scale quadcopters and handheld devices, respectively. We further explore this idea, developing an autonomy system that builds upon similar online forest inventory approaches~\cite{Freissmuth2024}.

Robotic platforms can support surveying tasks by freeing humans from collecting the data themselves~\cite{Anderson2013a}. In the context of aerial surveying, autonomous fixed-wing aircrafts \cite{Koh2012}, small-scale helicopters~\cite{Kellner2019} and more recently hexacopter platforms~\cite{Almeida2020} have been used for over-canopy data collection. As these platforms fly over the trees, most of the robotic challenges concern achieving maximum area coverage given energy \cite{Shah2020} and time constraints \cite{Schedl2021}.

Under-canopy data collection has shown more challenging, as it requires robots to operate in cluttered, unstructured spaces. For aerial platforms, this requires the use of smaller drones compared to the ones that can fly over canopy---with smaller payloads and battery size~\cite{Zhou2022}. Teleoperated small-scale drones have been used for more than ten years for surveying~\cite{Lin2011,Hyyppa2020} though autonomous operation has been recently demonstrated for forest inventory~\cite{Chen2020,Liu2022} and DNA sampling~\cite{Aucone2023}.

On the contrary, ground platforms are able to provide under-canopy measurements while carrying heavier sensing payloads, enabling longer surveys. Wheeled and tracked robot platforms have been used for forest data collection~\cite{Miettinen2007,Pierzchala2018,Tominaga2018} as well as the harvesting process \cite{Rossmann2009,Jelavic2021,Jelavic2022a}. However, their larger footprint and mass contribute to soil compaction damage due to trampling \cite{Fountas2010,Calleja-Huerta2023}.
In this work we aim to evaluate the performance of smaller platforms---up to \SI{50}{\kilo\gram}.

\subsection{Deployment of legged platforms in natural environments}
The recent advances in hardware and control of legged robots makes them an compelling alternative for ground-level surveying~\cite{Miki2022a}.

Legged robots have been used for teleoperated soil sampling~\cite{Wilson2021} as well as wildlife recording~\cite{Melo2023}. Autonomous operation in natural environments has also been demonstrated for search and rescue operations~\cite{Tranzatto2023}, load carrying tasks~\cite{Wooden2010, Bradley2015}, measurement of aeolian processes~\cite{Qian2017,Qian2019}, and planetary analog exploration~\cite{Arm2023}. 

Forestry applications have been less well unexplored, though their potential advantages (heavier payload carrying, lower soil impact) have been discussed in the past~\cite{Todd1985}. In this work, we aim to provide experimental demonstration of their capabilities for forest inventory applications.

\section{Autonomous Forest Inventory with Legged Robots}
\label{sec:method}

\begin{figure}[t]
  \centering
  \includegraphics[width=1\linewidth]{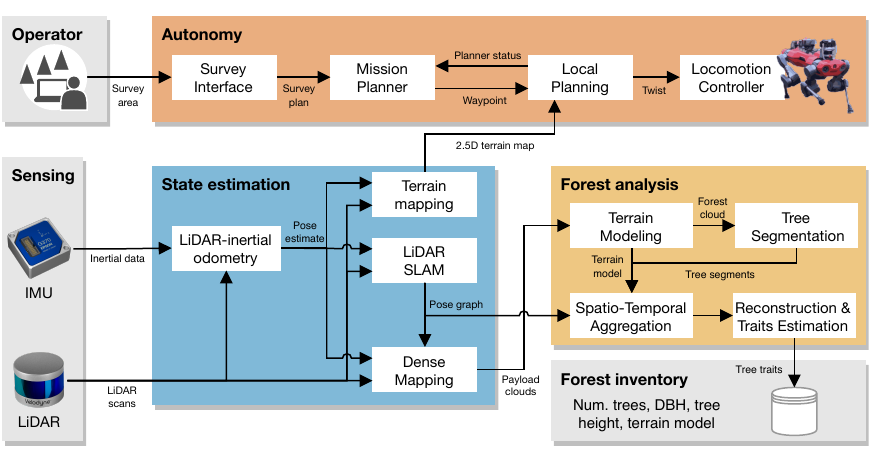}
  \caption{System overview. The \textbf{Autonomy system} executes the mission from a survey reference given by a human operator. \textbf{State estimation} provides a consistent scene representation, as well as dense clouds which are used for tree traits estimation. \textbf{Forest analysis} segments and estimates tree traits from point cloud data. The main output of the system is a forest inventory database with the main attributes of the surveyed forest.}
  \label{fig:alf-system-overview}
\end{figure}

\figref{fig:alf-system-overview} shows the system architecture of our solution, involving state estimation, legged autonomy, and forest analysis. The specific tasks and constituent modules are explained in the following sections.

%\the\textwidth
\subsection{State Estimation}
\label{sec:state-estimation}
Our state estimation approach involves four modules: (1) LiDAR-Inertial Odometry, (2) LiDAR SLAM, (3) dense mapping,  and (4) local terrain mapping, which provide representations for forest analysis as well as navigation.

\noindent\textbf{LiDAR-Inertial Odometry:} For odometry we use a LiDAR-inertial factor graph-based odometry solution, VILENS~\cite{Wisth2023}, which provides high-frequency pose and velocity estimates ($\sim$\SI{200}{\hertz}) in a fixed inertial frame. The system performs sensor fusion of relative pose displacements from LiDAR using the \gls{icp} algorithm~\cite{Besl1992} with preintegrated IMU measurements \cite{Forster2017}.

\noindent\textbf{LiDAR SLAM:} We use a pose graph LiDAR SLAM system~\cite{Ramezani2020a}, which incrementally builds a lightweight but consistent representation of the environment. Each graph node stores a LiDAR scan with the edge factors corresponding to incremental odometry estimates from LiDAR-Inertial odometry. We detect loop closures by checking the most recent scan against those linked to pose graph nodes within a radius of \SIrange{10}{15}{\meter} from the robot. The candidates are geometrically verified using \gls{icp}, and the successful candidates are added as additional factors in the pose graph. The graph is incrementally optimized using the \gls{isam}{2} algorithm~\cite{Kaess2012}.

\noindent\textbf{Dense Mapping:} As the SLAM system only stores sparse LiDAR scans within the pose graph, this map is not dense enough for further forestry analysis. We implemented a local sub-mapping module to accumulate all the incoming LiDAR scans and generate dense scans in the sensor frame, which we call \emph{data payloads}. The data payloads are gravity-aligned and published at a slower rate ($\sim$\SI{0.1}{\hertz}), accumulating clouds for about every \SI{20}{\meter} traveled. We process these data payload to carry out online tree traits analysis (see \secref{subsec:forest-analysis}).

\noindent\textbf{Local Terrain Mapping:} In addition to the large-scale mapping modules, we also carry out local terrain mapping~\cite{Fankhauser2018} and use its output in the local planner. We used a 2.5D grid-based representation with a resolution of \SI{4}{\centi\meter}.

\subsection{Autonomy System}
Our autonomy system is multiple levels: spanning from the Survey Interface (Level 3) down to the low-level locomotion control (Level 0).
% \mfallon{Its a bit weird to count downwards. meh, what can you do.} -> the human defines the mission and the lower layers execute

\noindent\textbf{Survey Interface (Level 3):} The user interface represents the highest level of planning of our system. The human operator directly interacts to deploy the robot. We implemented it as a graphical user interface (GUI) on RViz using interactive markers. %~\cite{Gossow2011}.
Given the initial pose of the robot, the GUI allows the operator to define a survey area and to initialize a survey plan, given by a boustrophedon decomposition, also known as lawn mower pattern~\cite{Choset2000}.

The survey plan is concretely defined as a sequence of equally-spaced 6 \gls{dof} waypoints, which enforce some minimum constraints to ensure that the \gls{slam} system is able to find loop closures during operation and can keep the trajectory and the map consistent. However, this autonomy level makes no assumptions about the traversability of the surveyed area and the reachability of the waypoints, which are delegated to the lower levels of the autonomy system.

The survey interface also provides mechanisms to interrupt the mission at any time, which triggers a safe stopping behavior through the lower levels. It also enables the operator to modify and resume the plan from any specified goal.

\begin{figure}[t]
  \includegraphics[width=\linewidth]{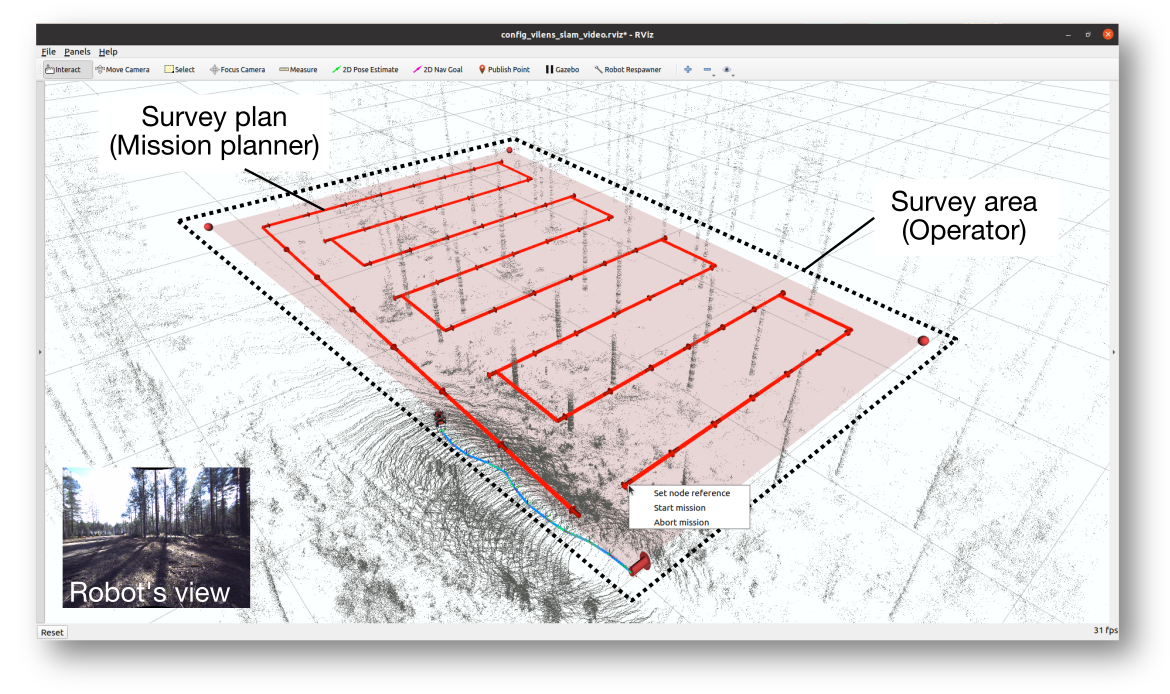}
  \caption{Survey Interface: GUI for the operator, implemented on RViz using interactive markers. The operator defines the area to be surveyed, while the survey plan is automatically proposed by the Mission Planner.}
  \label{fig:alf-mission-planning}
\end{figure}

\noindent\textbf{Mission Planner (Level 2):} The mission planner handles the execution of the survey plan by interacting with the local planner. In particular, it schedules the goals sent to the local planner and also processes any feedback obtained by it. For example, if the local planner reports a goal as unreachable, the mission planner chooses a new goal based on the survey plan's state.

\noindent\textbf{Local Planning (Level 1):} The local planning level aims to reach the proposed goals of the plan and checks its feasibility given the state of the local environment encoded in the terrain map. A traversability estimation system first processes the local terrain map and extracts geometric features to score the terrain. The traversability score $s_{\text{trav}}$ is specifically designed to enable navigation in the forest despite branches and twigs, though visual traversability approaches could also be used~\cite{Frey2023}.

The traversability representation is used by a reactive local planner~\cite{Mattamala2022}, which combines different vector fields for navigation. We extended the approach to compute a \gls{gdf} on a continuous cost-to-go $c_{\text{to-go}}$ obtained by the heuristic:
\begin{equation}
  c_{\text{to-go}} = w_{\text{trav}}\,(1.0 - s_{\text{trav}}) + w_{\text{unkn}}\,s_{\text{unkn}}
\end{equation}
where $s_{\text{unkn}}$ is a fixed score assigned to empty cells in the terrain map (unknown), and $w_{\text{trav}}$, $w_{\text{unkn}}$ are manually defined weights to leverage the contribution of the traversability score of known cells, as well as unknown ones.

The local planner measures the progress when trying to reach a waypoint, in order to detect potential situations were it is unfeasible (e.g., if the proposed waypoint is in the middle of bushes).
If a goal is determined unreachable, the local planner reports this situation to the mission planner, triggering a replanning of the next goal.

\noindent\textbf{Locomotion Controller (Level 0):} The output of the local planner is a 3 \gls{dof} velocity command---linear $(x,y)$, angular (yaw)---, which is executed by a low-level reinforcement learning-based locomotion controller~\cite{Miki2022a}. The controller specifies a nominal walking wait for flat ground, but it relies on the local terrain map to negotiate uneven, rigid terrain. We used a pre-trained policy and did not implement any specific modifications to the architecture or training environment to operate in the forest.

\subsection{Online Forest Analysis}
\label{subsec:forest-analysis}
Our system integrates a state estimation-driven forest inventory pipeline, presented in Frei{\ss}muth \etal{}~\cite{Freissmuth2024}. The approach enables near real-time tree segmentation and reconstruction during the mission execution. The main steps are explained as follows, and please refer to the original work for more technical details of the implementation.

\noindent\textbf{Terrain Modeling:} The input to the online forest analysis step are the data payloads at \SI{0.1}{\hertz}. For each payload, we extract a local terrain model using the \emph{cloth simulation filtering} method by Zhang \etal{}~\cite{Zhang2016b}. This enables us to segment the points belonging to the terrain from the potential tree points.

\noindent\textbf{Tree Segmentation:} To segment trees we use a method inspired by Cabo \etal{}~\cite{Cabo2018}. We first normalize the payload cloud z-values using the terrain model, and only keep the points in a slice where we expect no tree foliage. We then use a Voronoi-based clustering to obtain tree cluster candidates. We refine the points associated to each tree by fitting a cylinder model and only keeping the points within a distance from the cylinder's surface. The filtered tree points per tree candidates are finally de-normalized by height to obtain tree candidate clouds in the original reference frame.

\noindent\textbf{Spatio-temporal Aggregation:} Our system is designed to incrementally build a forest inventory, effectively improving the tree estimates and terrain model as the robot autonomously collects more data. We achieve this by exploiting the pose graph representation of the LiDAR-SLAM system, by attaching local terrain models and tree instances clouds to it, and updating them with new data after adding new poses or loop closure updates. Specifically, the global terrain model is updated by locally averaging the points around the latest terrain map using a weighted average. The tree instances clouds are managed accordingly to decide if new instances correspond to an existing tree (due to proximity) or a novel instance. We additionally check coverage conditions to ensure that high points are captured as well as even coverage from different viewpoints.

\noindent\textbf{Tree Reconstruction and Trait Estimation:} Lastly, we estimate the traits required by the forest inventory. Particularly, we first aim to \emph{reconstruct} the tree by fitting circles along the stem, and then reconstructing the trunk as a collection of oblique cone frustums. The reconstructions are then used to recover geometric traits such as the \gls{dbh} and tree height. The traits are reported to the operator and also exported as a \emph{marteloscope}---a 2D representation of the forest.

\section{Experiments}
\label{sec:experiments}

\begin{table*}[t]
    \centering
    \footnotesize{
    \begin{threeparttable}
    \setlength{\tabcolsep}{3pt}
\begin{tabular}{C{1.5cm} C{1.5cm} C{0.5cm} C{1.2cm} C{1.5cm} | C{1.5cm} C{1.5cm} | C{1cm} C{1.2cm} C{1cm} C{0.8cm} C{0.8cm} C{0.8cm}}
\toprule
\multicolumn{4}{c}{\textbf{Mission}} & \multicolumn{2}{c}{\textbf{Robot setup}} & \multicolumn{6}{c}{\textbf{Metrics}}
\\
\textbf{Campaign} & \textbf{Date} & \textbf{ID} & \textbf{Area} [$m^2$] & \textbf{Goals} & \textbf{Hardware} & \textbf{Software} & \textbf{Mission time} [$s$] & \textbf{Distance traveled} [$m$] & \textbf{Area covered} [ha] & \textbf{Interv.} [$\#$] & \textbf{MDBI} [$m$] & \textbf{MTBI} [$m$]
\\ \midrule
\multirow{5}{1cm}{\centering Evo, Finland} & 2023-05-03 & M1 & $40 \times 25$ & \multirow{5}{1.5cm}{\centering Autonomy system} & \multirow{5}{1.5cm}{\centering ANYmal C, Velodyne VLP-16} & \multirow{5}{1.5cm}{\centering CompSLAM odometry} & 575.6 & 270.3   & 0.33 & 2 & 84.8 & 176.1 \\
  & 2023-05-03 & M2 & $40 \times 25$ &   &   &   & 432.4  & 233.6 & 0.32 & 0 & 233.6 &  432.4 \\
  & 2023-05-04 & M3 & $40 \times 35$ &   &   &   & 816.8$^\star$  & 301.1  & 0.37 & 7 & 30.5 &  72.2 \\
  & 2023-05-04 & M4 &  $35 \times 35$ &   &   &   & 988.4  & 336.6  & 0.37 & 7 & 39.4 & 114.1  \\
  & 2023-05-05 & M5 &  $70 \times 25$ &   &   &   & 1275.5 & 609.7  & 0.64 & 10 & 51.5 & 99.5  \\ \midrule
Wytham Woods, UK & 2023-10-06 & M6 &  $20 \times 20$ & Autonomy system, State estimation & ANYmal C, Velodyne VLP-16 & VILENS odometry, VILENS-SLAM & 436.9 & 215.0  & 0.29 & 2 & 69.9 & 136.4 \\ \midrule
Forest of Dean, UK & 2023-02-19 & M7 & $125 \times 30$ & State estimation, Forest inventory& ANYmal D, Hesai QT64 & VILENS odometry, VILENS-SLAM & 1283.5 & 665.5  & 0.96 & 8 & 65.9 &  127.7 \\ \bottomrule
\end{tabular}
    % Table footnotes
	\begin{tablenotes}\footnotesize
		\item[$\star$] This mission was manually interrupted.
	\end{tablenotes}
	\end{threeparttable}
    }
    \caption{Missions summary. We report the specifications of each mission executed across three campaigns in Finland and the UK, along with the robot setup used, as well as the main metrics reported on each mission, such as the \glsfirst{mdbi} and \gls{mtbi}.}
    \label{tab:mission-summary}
\end{table*}

\begin{figure*}[t]
    \centering
    % \missingfigure{Show real pictures of the robot deployed}
     \includegraphics[width=\linewidth]{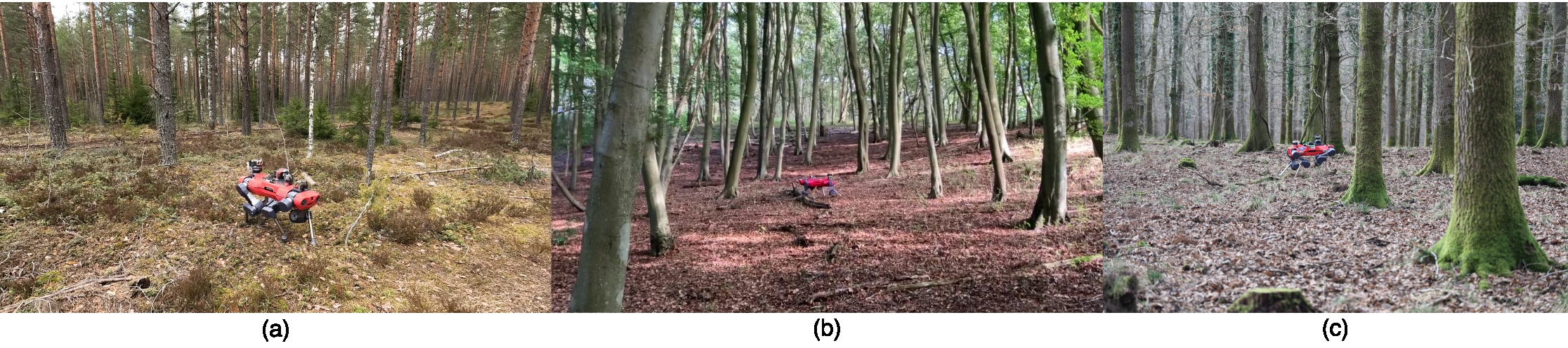}
    \caption{\small Illustrative examples of the robot deployments during the different campaigns. (a) Coniferous forest in Evo, Finland. (b) Mixed forest in Wytham Woods, UK. (c) Oak forest in the Forest of Dean, UK.}
    \label{fig:real-examples}
\end{figure*}

\subsection{Campaign description}
We executed three field campaigns in different forests in Finland and the UK, which are summarized in \tabref{tab:mission-summary}. Five missions (M1-M5) were executed in conifer forests in Evo, Finland in May 2023. The other two were ran in the UK, in mixed (M6, at Wytham Woods, in October 2023) and oak forests (M7, at the Forest of Dean, in February 2024). While the two former had the objective of testing the autonomy system, the latter also focused on the online forest analysis integration. \figref{fig:real-examples} illustrates some examples of the robot deployed in the field.

\subsection{Deployment methodology}
All the missions followed the same procedure. A \emph{robot operator} was in charge of defining the survey mission and supervising the overall execution. A \emph{safety operator}, carrying a remote controller to overrule the autonomy system, followed the robot at a safe distance during the mission to avoid undesired situations (e.g. collisions and \emph{cul-de-sacs}).

\begin{figure}[t]
    \centering
    \includegraphics[width=\linewidth]{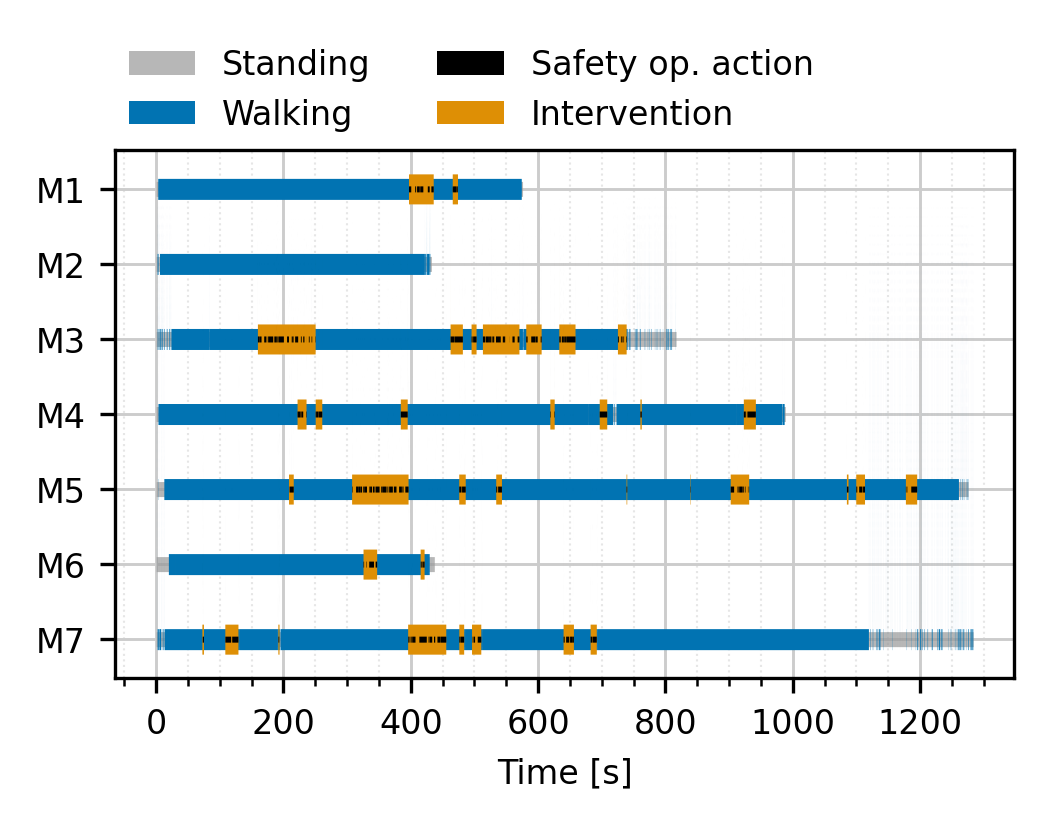}
    \caption{\small Summary of the seven survey missions executed across the three campaigns. We illustrate the periods where the robots operated fully autonomously, and where they required manual interventions from the safety operator.}
    \label{fig:mission-summary}
\end{figure}

We logged all the signals from the robot and the autonomy system, as well as the actions executed by the safety operator. The safety interventions were the main signal we used for evaluating the performance of the autonomy system, as it enables the use of metrics such as the \gls{mdbi} and \gls{mtbi}, as done in autonomous driving literature~\cite{Paz2020}. However, instead of computing the metrics using the total distance traveled over the number of interventions reported, we determined the average by measuring the actual time and distance intervals when the robot was effectively walking autonomously. We call these measurements \emph{autonomy segments}, shown in blue in the summary plot (\figref{fig:mission-summary}).

\subsection{Evo Campaign}
Our first campaign, in Finland, had the goal of evaluating the proposed autonomy system in forest environments. This was executed with an ANYmal C platform, using the Velodyne VLP-16 (\SI{30}{\degree} \gls{fov}) as the main environmental sensor. We used a state estimation setup derived from the CERBERUS SubT stack~\cite{Tranzatto2022}, which used CompSLAM~\cite{Khattak2020} as the main odometry system. 
CompSLAM does not perform online pose graph optimization and loop-closure detection to ensure consistency.

\begin{figure}[t]
    \centering
    \includegraphics[width=0.45\linewidth, trim={0cm 2cm 0cm 2cm},clip]{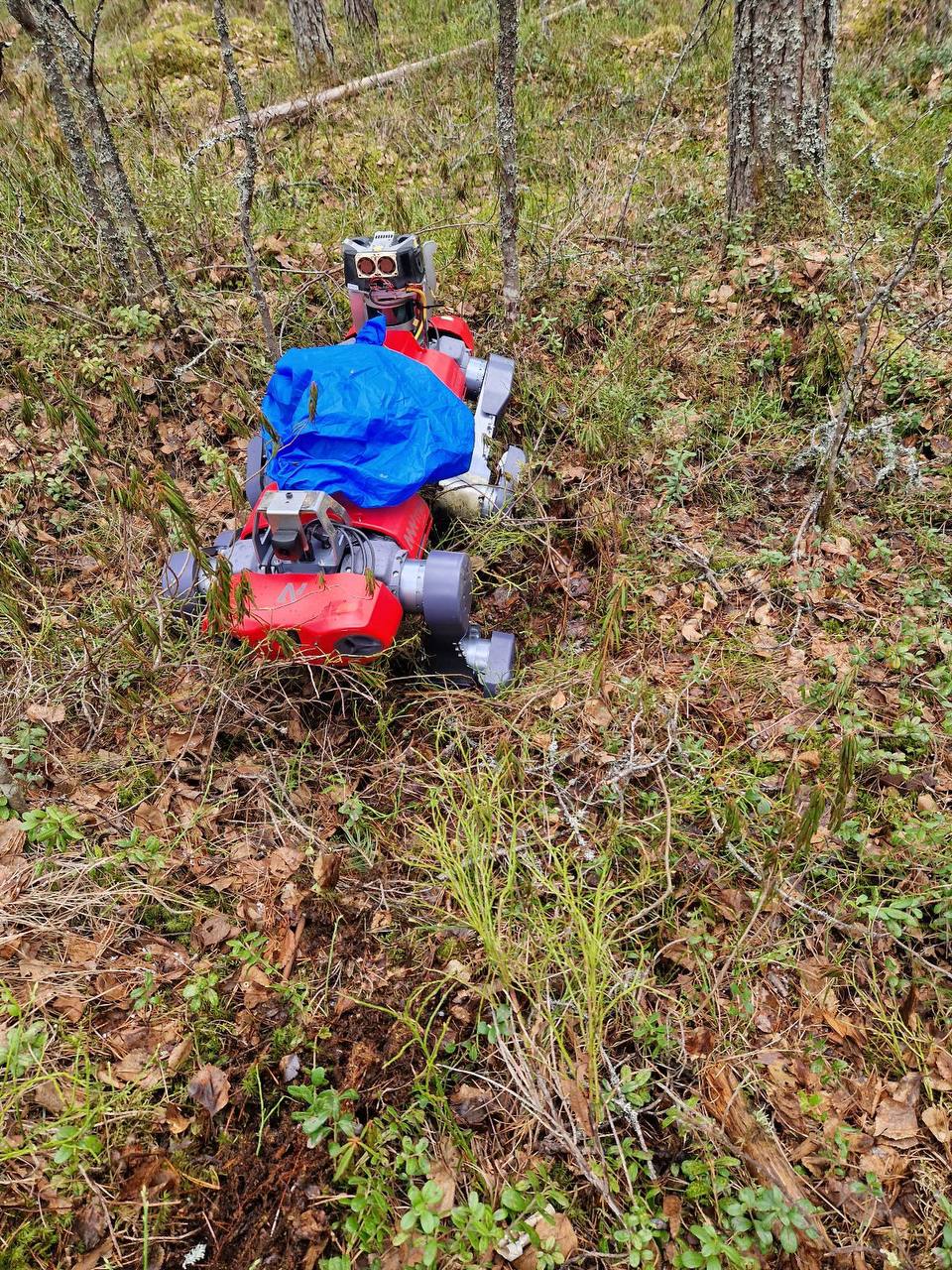}
    \includegraphics[width=0.45\linewidth, trim={0cm 2cm 0cm 2cm},clip]{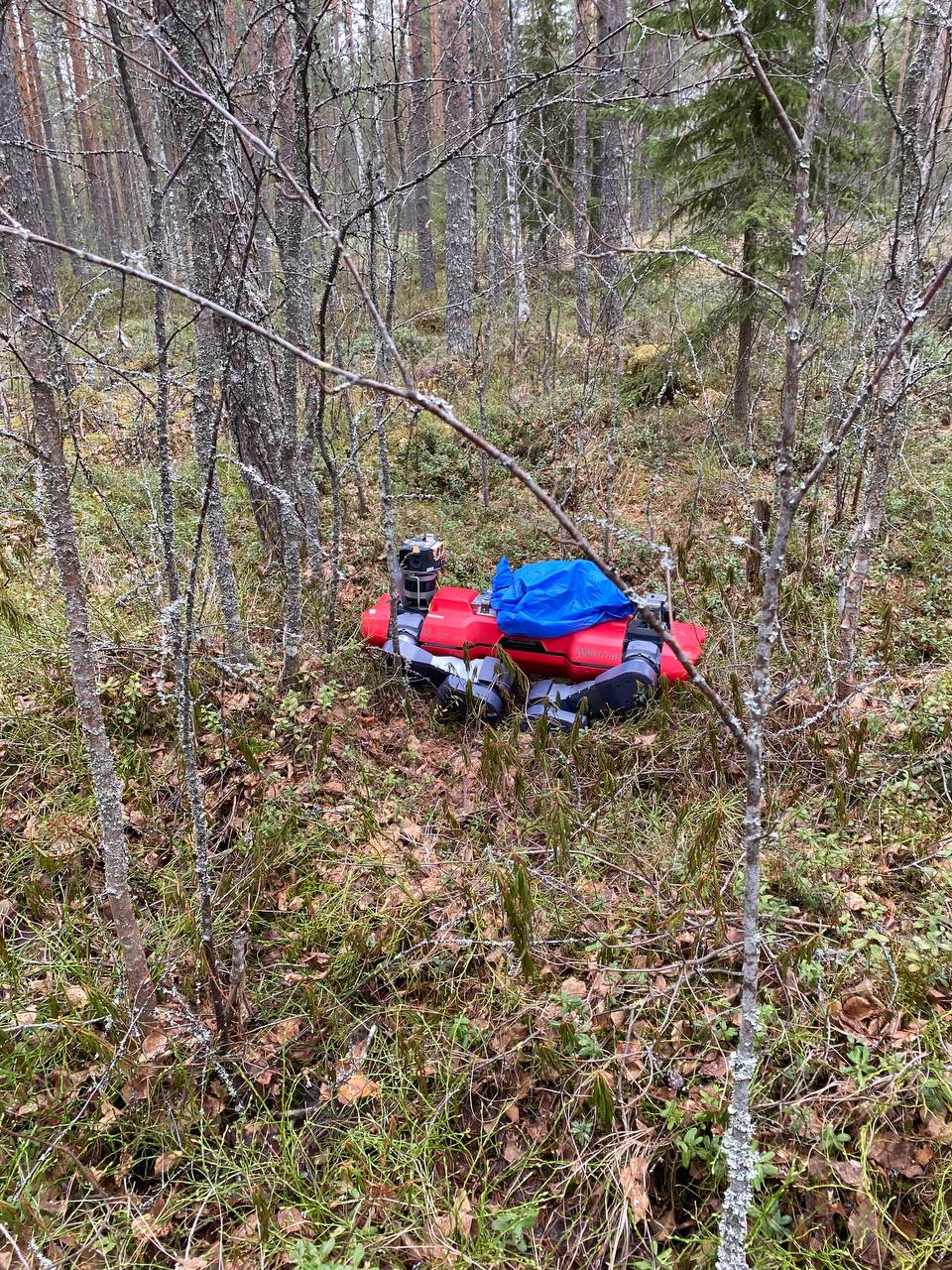}
    \caption{\small Failure case in mission M3 in Evo, Finland. The robot was trapped due to damp terrain, which was not detected by the onboard perception systems.}
    \label{fig:failure}
\end{figure}

\figref{fig:mission-summary} shows the autonomy performance reported in these missions (M1-M5), while \figref{fig:real-examples} (a) shows an example from M2. Missions M1 and M2 were executed in the same plots, obtaining consistent results. Mission M3 was executed in an area nearby, but with more challenging terrain, which required to interrupt the mission due to the robot getting trapped in a damp area (\figref{fig:failure}). Missions M4 and M5 were executed in different areas of the forest, but achieving similar area coverage per time unit ($~$\SI{1.5}{\hectare}), as shown in \tabref{tab:mission-summary}. In M5, we additionally observed severe tracking problems with the reference survey path due to drift in the state estimation system, which were more evident in this longer sequence (\SI{600}{\meter} long, compared to \SI{300}{\meter} set for the other missions).

\begin{figure}[h]
    \centering
    \includegraphics[width=\linewidth]{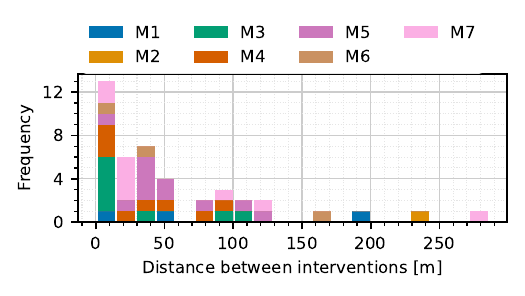}
    \includegraphics[width=\linewidth]{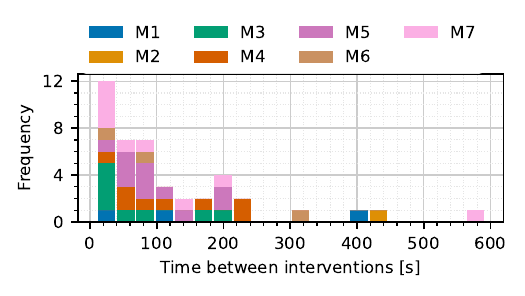}
    \caption{\small Distribution of distance and time between interventions for all missions. \emph{Top:} Distance between interventions. \emph{Bottom:} Time between interventions.}
    \label{fig:between-interventions}
\end{figure}

The missions lasted between \SIrange{10}{20}{\minute}, with the robot walking up to \SI{600}{\meter} on the longest missions. About the autonomy performance, while we report the \gls{mdbi} and \gls{mtbi} in \tabref{tab:mission-summary}, we found the fine grained analysis of the distribution of autonomy segments a more informative metric. This is shown in \figref{fig:between-interventions}.
With exception of mission M2, which was fully autonomous, in most of the other missions we reported a higher density of short autonomy segments ($\sim$\SI{10}{\meter} or $\sim$\SI{50}{\second}). Many cases were due to short, frequent `pushes' performed by the safety operator to move the local planner solution out of local minima, or driving the robot around a dead end.

\begin{figure}[h]
    \centering
    \includegraphics[width=\linewidth]{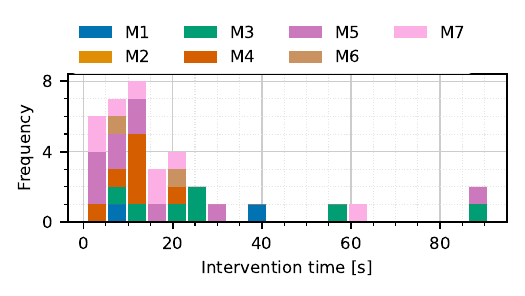}
    \caption{\small Duration of the interventions reported for each mission. Most of the interventions reported were short---\SI{20}{\second} on average.}
    \label{fig:time-interventions}
\end{figure}

We additionally reported how long the interventions took on each mission (\figref{fig:time-interventions}). We observed that most of the interventions were $\sim$\SI{15}{\second}, which is within the time required for the safety operator to get the robot out of dead ends that the mission and local planner interaction was not able to solve automatically. Only in specific cases---M3 and M5---, where the robot was deployed in considerably more challenging environments (damp terrain, short trees, higher density of bushes) we required longer interventions ($>$\SI{1}{\minute}) to move the robot to safer areas before continuing the mission.

\subsection{Wytham Woods Campaign}
The second campaign was executed in the UK, in a mixed forest plot in Wytham Woods, near Oxford. The same ANYmal C platform and LiDAR setup was used in this campaign, as the goal was to improve the state estimation solution problems reported in the Evo campaign. For this, we transitioned to the proposed state estimation solution, which explicitly relied on a LiDAR-inertial odometry system along with an online pose graph SLAM to ensure consistency.

\figref{fig:real-examples} (b) shows the testing environment, which was a clear sloped plot, with a few loose branches and twigs. We set a smaller \SI{20}{\meter}~$\times$~\SI{20}{\meter} survey area, which corresponded to \SI{0.29}{\hectare} covered for an effective \SI{15}{\meter} LiDAR range. The reported coverage rate was consistent with the results obtained in the Evo campaign, above \SI{1.5}{\hectare\per\hour}. Only two short interventions were required (see \figref{fig:time-interventions}, M6) to avoid falls due to loose branches, which trapped the robot's legs.

\subsection{Forest of Dean Campaign}
The last campaign, executed in the Forest of Dean in the UK, was used to test the integration of the forest analysis system towards online forest inventory. Additionally, we changed the robot platform from ANYmal~C to ANYmal~D---which has longer shanks, though the locomotion policy does not exploit this fact explicitly. We also improved the sensor payload, where we used a sensing unit with a dedicated computer and a Hesai QT64 LiDAR, which as a larger vertical \gls{fov} (\SI{104}{\degree}), enabling better coverage of the full trees. The state estimation and autonomy were left unchanged apart from the sensor configuration, as they were validated in the previous campaigns.

\begin{figure*}[t]
    \centering
    \includegraphics[width=\textwidth]{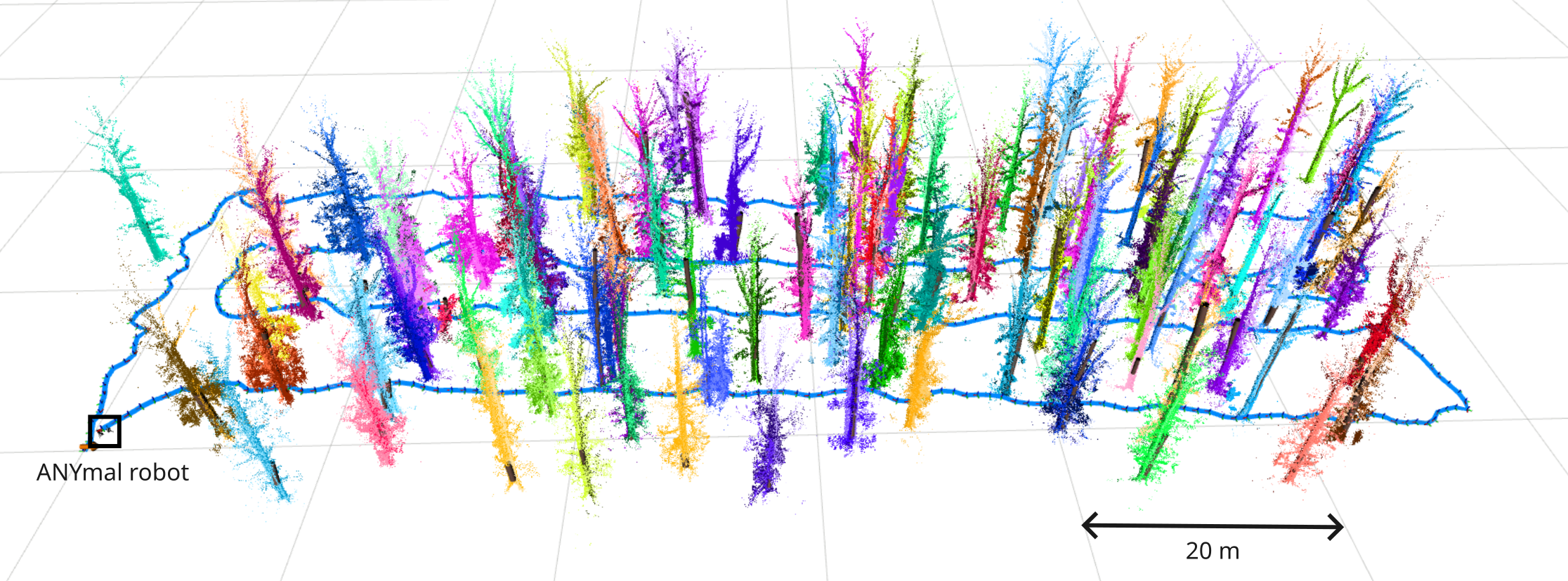}
    \caption{\small Illustrative example of the output produced by our system after an autonomous mission in the Forest of Dean, UK. The robot surveyed \SI{0.96}{\hectare} in \SI{20}{\minute}, and around $100$ trees were segmented during online operation.}
    \label{fig:forest-dean}
\end{figure*}

\begin{figure*}[t]
    \centering
    \includegraphics[width=0.9\textwidth]{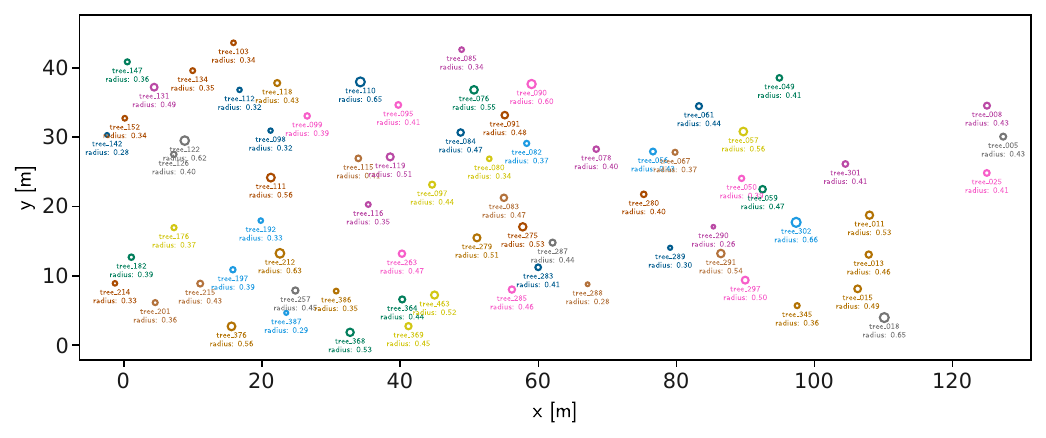}
    \caption{\small Marteloscope produced after the deployment in the forest of Dean, indicating the tree positions and the \gls{dbh} of the successfully reconstructed trees.}
    \label{fig:forest-dean-marteloscope}
\end{figure*}

For the specific mission we executed, denoted M7, we specified a \SI{125}{\meter}~$\times$~\SI{30}{\meter} survey area, which effectively corresponded to a \SI{0.96}{\hectare} plot. The robot completed the mission in approximately \SI{20}{\minute}, segmenting up to $100$ trees, as shown in \figref{fig:forest-dean}. \figref{fig:forest-dean-marteloscope} shows the marteloscope, which summarizes the tree positions and \gls{dbh} reported for the trees successfully reconstructed by our proposed online forest analysis pipeline.

\section{Current Challenges and Future Work}
\label{sec:limitations}
Our deployments so far have allowed us to continuously refine and improve our proposed forest inventory solution, from a software and hardware perspective. Our experiments to date indicate that, given the robot capabilities and the sensing setup, our solution should enable the autonomous inventory at \SI{2}{\hectare\per\hour}. This is a feasible bound given the battery specifications of the ANYmal platform, assuming fairly clean forest areas without challenging ground vegetation. Nevertheless, we still observe open challenges to address to improve the performance of our systems towards autonomous inventory and monitoring.

% Added by Nived: Feel free to change it.
\noindent\textbf{Autonomy performance:} Firstly, we aim to improve the performance reported through the \gls{mdbi} and \gls{mtbi} metrics. Across the missions we did not observe a clear trend of improvement, partly caused by the changes in the constituent components but also due to measuring the metrics in completely different environments. To improve performance, we identified potential improvements in the autonomy system. Particularly by designing new re-planning strategies to detect and navigate around untraversable areas effectively, identifying and avoiding dead ends, as well as maximizing loop closure corrections.

\noindent\textbf{Locomotion and navigation:} In spite of using a robust locomotion controller, we observed that further improvements are still required to negotiate branches and twigs in a safer manner, to avoid potential falls and failures due to trapped legs. This could be solved by low-level strategies to vary the step size, or by identifying such areas with the onboard perception systems, effectively guiding the robot through longer-though-safer paths.

\noindent\textbf{Forestry tasks integration:} Lastly, we identified further improvements can be achieved for the real-world usability of our solution. Practical issues such as improving long-range communication in the forest, could enable better feedback of the operation, and provide live reporting of the estimated traits for the foresters. Additionally, studying the impact of legged locomotion on the forest soil, as well as practical comparisons to alternative inventory approaches could further contribute to the adoption of these platforms for forestry.

\section{Conclusion}
\label{sec:conclusion}
In this work, we presented ongoing efforts to develop an autonomous forest inventory system with small-scale legged platforms. Through seven different missions in forests in Finland and the UK, we have demonstrated the feasibility of using legged platforms for these tasks, providing a systematic way to survey and study natural environments. We have also presented open challenges, which we identified during our deployments. While some of them will be improved for future field deployments, they also present research questions that can be of interest to the field and legged robotics communities.

\section{Acknowledgments}
This work is supported by EU Horizon 2020 (101070405 DigiForest) as well as a Royal Society University Research Fellowship. We acknowledge PreFor Oy for organizing the Evo campaign, as well as Forest Research UK for arranging the Forest of Dean campaign.

% Bibliography
\balance
\bibliographystyle{IEEEtran}
\bibliography{references.bib}

\end{document}